# A Comparative Study of Pre-trained CNNs and GRU-Based Attention for Image Caption Generation

Rashid Khan*[1, 2]  Bingding Huang*,[2], Haseeb Hassan[3], Asim Zaman [3] and Zhongfu Ye[1]

[1]University of Science and Technology of China, Hefei, 230026, Anhui, China
[2]College of Big Data and Internet, Shenzhen Technology University, Shenzhen, 518188
[3]College of Health Sciences and Environmental Engineering Shenzhen Technology University, Shenzhen, China）

Corresponding authors: *Corresponding author: Rashid Khan (rashidkhan@sztu.edu.cn) and, Bingding Huang (huangbingding@sztu.edu.cn)

**Abstract**
Image captioning is a challenging task that involves generating a textual description for an image using both computer vision and natural language processing techniques. In this paper, we propose a deep neural framework for image caption generation using a GRU-based attention mechanism. Our approach employs multiple pre-trained convolutional neural networks as the encoder to extract features from the image and a GRU-based language model as the decoder to generate descriptive sentences. To improve performance, we integrate the Bahdanau attention model with the GRU decoder to enable learning to focus on specific parts of the image. We evaluate our approach using the MSCOCO and Flickr30k datasets and show that it achieves competitive scores compared to state-of-the-art methods. Our proposed framework can bridge the gap between computer vision and natural language and can be extended to specific domains..

## 1. Introduction

Image caption generation is an essential problem in computer vision (CV) and natural language processing(NLP) that aims to generate descriptive sentences for images. Recent research on caption generation has focused on the encoder-decoder framework, where a neural network architecture is employed as the encoder to extract features from the source image, and a language model is used as the decoder to generate the target sentence. While this approach has shown promising results, attention mechanisms are necessary to evaluate the importance of the encoder's hidden states. However, sequential attention mechanisms lack global modeling capabilities and require a review network to generate compact and abstractive annotation vectors..

On the other hand, NLP has spawned a slew of applications ranging from primary text classification to fully automated natural language chatbots. In the Deep Learning domain, it has been a critical and fundamental challenge. Image captioning (IC) has a wide range of applications, including (i) transcribing scenes for people who are blind [1,2], (ii) classifying videos and photographs based on various situations [3], (iii) image-based search engines for better results, [4] (iv) visual question answering [5], and (v) context comprehension [6].

To generate the corresponding sentence for a given image, the latest research on caption generation, such as image captioning [7], relies on an encoder-decoder framework. Different neural network architectures are employed as the encoder due to the other behavior and features of the source, like convolutional neural networks (CNNs) for images and recurrent neural networks (RNNs) for sequential data, including source code and natural language. The attention technique evaluates the significance of the encoder's hidden states based on all previously generated words in the target sentence at each time step. The attention mechanism, on the other hand, works in a sequential manner and lacks global modeling capabilities. A review network [8] was proposed to overcome this flaw, with review steps located here between the encoder and the decoder. As a result, more compact, abstractive, and global annotation vectors are generated, which have been shown to assist the sentence generation process further in. We demonstrate a CNN and GRU-based attention mechanism for automatic image captioning in this research work. The suggested framework was designed with one encoder-decoder framework. As the encoder, we used various pre-trained convolutional neural networks to encode an image into a feature vector as graphical characteristics. Subsequently, to create the descriptive sentence, a language model called GRU was chosen as the decoder. However, we combined the Bahdanau attention model with GRU to allow learning to be focused on a specific portion of the image in order to improve performance.

The following are the main contributions of our paper:

1. For image caption generation, we examined the encoder-decoder framework. The ENCODER would use a pre-trained Convolutional Neural Network (CNN) to encode the image, and the DECODER would use a Recurrent Neural Network (RNN) to create each word of the caption iteratively.
2. The model's performance was compared to four pre-trained CNNs: InceptionV3, DenseNet169, ResNet101, and VGG16. We employ the GRU with soft attention as the decoder, which effectively focuses the attention over a specific part of an image to predict the next sentences.
3. In our image captioning approach, we apply an attention mechanism that can focus on the essential elements of the image and define fine-grained captions. Finally, we utilize the open-source MSCOCO and Filker30k datasets to quantitatively validate the research paper's utility in image caption generation.

The rest of this paper is structured in the following manner. Section 2 starts with the previous work and an overview of our framework. Following that, each module of our technique is presented in depth. In section 3, we describe our proposed framework in detail. The results of our approach are shown in Section 4 as an evaluation of the suggested methodology. In section 5, go through the future directions and how they may be further explored.

## 2. Related Work

In related work, we enhance relevant information on a prior study on image caption generation and attention. Several approaches for generating image descriptions have recently been presented. Automatic image captioning generation has emerged as a promising research area in recent years because of advances in deep neural network models for CV and NLP. In general, there are three types of image captioning modeling techniques: neural-based approaches [10-12], attention-based strategies [13-16], and RL-based methods framework [17, 18]. Attention-based approaches have recently gained popularity and are more successful than neural-based methods. When guessing each word in the caption, attention-based techniques focus on specific locations in the image.

Deep neural networks (DNNs) were initially proposed for caption generation [19]. They suggested extracting characteristics from images using convolutional neural networks (CNNs) to produce captions. A popular captioning technique involves integrating CNN and RNN, with CNN extracting image features and RNN framework generating the language model [20], for example, presenting an end-to-end network made up of a CNN and an RNN. Given the CNN feature of the training image at the starting time step, the model is trained to maximize the likelihood of the target sentence. The image's CNN feature is provided in the multimodal layer after the recurrent layer rather than at the beginning of the proposed m-RNN model [21]. Some are comparable instances of similar work [22, 23] that utilize CNN and RNN to generate descriptions.

Visual attention is an efficient image caption generation approach [24],[25]. When developing the target language, these attention-based captioning models may learn where to focus on the image. They may understand the distribution of spatial attention during the last convolutional layer of the CNN [26], or they learn the distribution of semantic attention from visual characteristics known from social media images [27]. Whereas these methodologies demonstrate the efficiency of the attention mechanism, they do not investigate the contextual information in the encoding sequence. Our attention layer is distinct in that it is structured in sequential order, with each hidden state of an encoding stage contributing to the formation of decoding words.

An Attention mechanism can be used to improve the contextual aspect of natural language sequences. The use of attention to describing image content is consistent with human understanding [24]. The evaluation matrix and the accuracy of attention to imaging correlate significantly. Even still, the measure to which attention accuracy is congruent with human perceptions needs to be increased [28]. The attention area captioning model comprises image regions, word captions, and the NLP natural language framework (RNN). The MS COCO dataset is typically used to test the trained system [29].

In artificial intelligence, developing a caption that accurately represents an image is essential [30]. One of the procedures in image captioning is extracting coherent characteristics of an image using an image-based framework. The extracted features of the image-based model are used to describe an image in NL. We suggested creating an image captioning model that employs a GRU with a soft attention decoder to predict future sentences by selectively focusing attention on a specific part of an image. We used cutting-edge architecture to evaluate the model's performance to that of four pre-trained CNNs: Inception V3, DenseNet169, ResNet101, and VGG16. The Attention layer is used to make the image's caption more sensible.

## 3. Caption Generation Using GRU-Based Attention Network

Extracting visual information and expressing it in a grammatically correct natural language sentence are the two main components of automatically generating natural language sentences that describe an image. Figure 1 shows a simple Encoder-Decoder deep learning-based captioning infrastructure for image captioning.

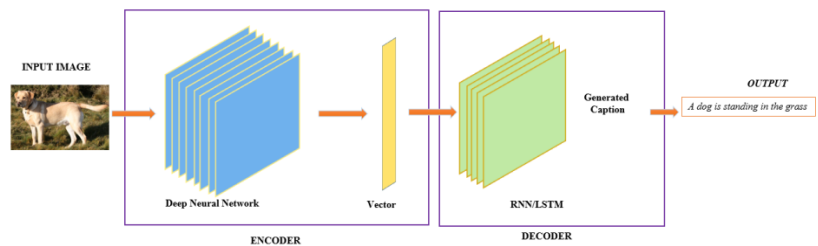

**Fig. 1.** This is an image captioning model's overall encoder-decoder structure. The image is encoded into a feature vector by a deep neural network. The language model uses the input vector to construct a sentence that defines the image

As previously stated, our approach is significantly influenced by the work of Xu et al. [31]. Their performance has improved as a result of our efforts. The proposed approach generates a caption encoded as a sequence of encoded words from a single input image. The CNN-Encoder, attention mechanism, and RNN-decoder are the three major components of our proposed framework. They work in sequence, with the images with captions as the CNN-encoder's input. The attention technique and the LSTM work in conjunction to generate captions for the input image and the output of this is passed to them. The objects and features in the image are retrieved using a convolutional neural network, and then we require a network to construct a meaningful sentence using our information.

$$q = \{q_1, q_2, \ldots, q_c\}, q_i \in R^K \tag{1}$$

A pre-trained CNN as encoder extracts features from images, an attention mechanism weights the image features, and an RNN as decoder provides captions to represent the weighted image features. The overall graphical representation of our framework is shown in Figure 2. Where $K$ denotes the vocabulary size and $c$ represents the caption length.

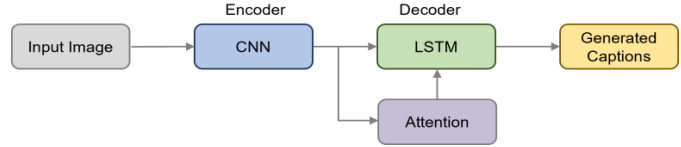

**Fig. 2.** The graphical representation of the framework

### 3.1. Convolutional Neural Network (Encoder)

A CNN pre-trained for an image classification task is commonly used as the encoder in the encoder-decoder framework for image captioning to extract the global representation and sub-region representations of the input image. A fully connected layer's output is usually the global representation, while a convolutional layer's output is usually the sub-region representation. As an encoder, we employed a convolutional neural network. A common feedforward neural network is a convolutional neural network. Convolutions and element-wise summations are performed by each layer as part of an affine operator. A visual representation is denoted as Figure 3, which illustrates an approach based on the architecture proposal of Xu et al. [31].

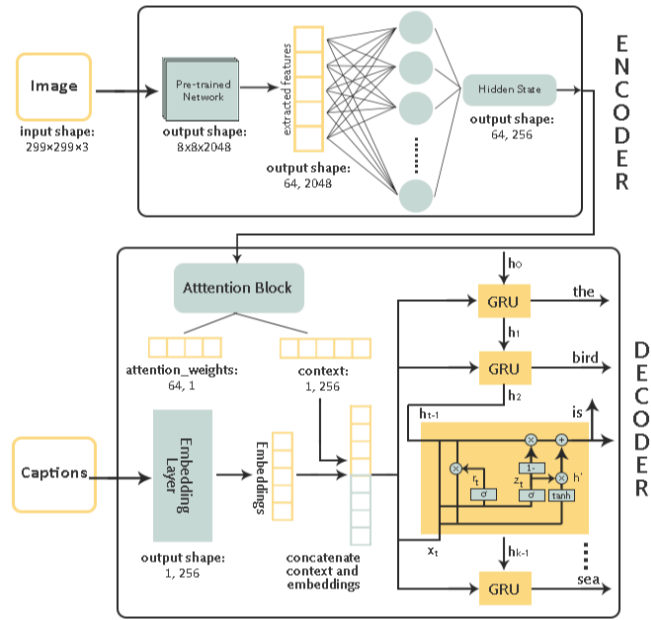

Fig. 3. Figure showing visual representation of approach. It is based on Xu et al [31] architecture proposal. The input image is sent to the encoder, which then transfers the output of the convolutional layers to the GRU with attention mechanism

$$f(x) = \sum_{k=1}^{K} g_k \times x + b_k \tag{2}$$

To convert images into a fixed-size vector, we can use the convolutional layers of pre-trained neural networks like Inception, DenseNet169, ResNet101, and VGG16. These networks are trained on large datasets and have learned to extract relevant features from images. Here's how we use each of the models mentioned to extract features from images:

*Inception:* ResNet-101 is a 101-layer deep convolutional neural network. The network learns rich feature representations for a wide range of images because of the many layers. The network's image input size is $224 \times 224$ pixels. It generates a $7 \times 7 \times 2048$ feature vector.

*DenseNet169:* DenseNet-169 is a 169-layer model that has been pre-trained. It has fewer parameters than other approaches, and the architecture effectively tackles the vanishing gradient problem. The network accepts a 224-by-224 image as input and outputs a $7 \times 7 \times 1664$ feature vector.

*ResNet101:* ResNet-101 is a 101-layer deep convolutional neural network. The network learns rich feature representations for a wide range of images because of the many layers. The network's image input size is in $224 \times 224$ pixels. It generates a $7 \times 7 \times 2048$ feature vector.

*VGG16:* The VGG pre-trained model was released by researchers from the Oxford Visual Geometry Group, who participated in the ILSVRC challenge. By default, the model expects color input photos to be rescaled to $224 \times 224$ squares. It generates a $7 \times 7 \times 512$ feature vector.

These networks were trained to classify 1000 different classes of images using the ImageNet dataset. Our goal is not to organize the image but to obtain a fixed-length informative vector for each image. As a result, we eliminated the model's last softmax layer and extracted a fixed-length vector for each image. These extracted characteristics are sent to the RNN, which creates the hidden state by using a fully connected layer.

### 3.2. Recurrent Neural Networks

While CNNs perform signal processing, they tend to describe patterns in sequences. The outputs of RNNs are fed back into the input, similar to feedforward neural networks. Throughout iterations of this feedback loop, the networks preserve a hidden state that allows them to change their behavior. RNNs are regarded as state-of-the-art in machine translation and other jobs involving text generation because they acquire human grammatical patterns quickly. In this research paper, we use a GRU, a sophisticated variant of the RNN approach that produces a caption by creating one word at each time step based on a context vector, the initial hidden state, and previously made words. The most likely description of an image is obtained in the encoder-decoder approach by maximizing the log-likelihood function of the expression $E$, considering the related image $I$ and the model parameters $\theta$.

$$\theta^* = \arg\max \sum (I, E) \log p(S | I; \theta) \tag{3}$$

Since $S$ it can represent any sentence length, a chain rule is commonly employed to characterize the joint probability over $E_1, E_2, \ldots, E_N$.

$$\log p(E|I) = \sum_{t=0}^{N} \log p(E_t | I, E_0, \ldots, E_{t-1}) \qquad (4)$$

For the sake of clarity, the dependency on $\theta$ is excluded. The network training is represented by the pair of (*E, I*), and we use the *Adam optimizer* to maximize the sum of the log-likelihood functions across the full training set. A recurrent neural network is used to represent the likelihood $\log p(E_t | I, E_0, E_1 \cdots, E_{t-1})$ where there are variable numbers of words that we define up to $t-1$. Using a fixed-sized hidden vector, RNN gives a seamless technique to execute conditioning on prior variables.

$$p(E_t | I, E_0, E_1, \cdots, E_{t-1}) \approx p(E_t | I, h_t) \qquad (5)$$

As a result, at step $t$, a simple vector replaces the complex conditioning on a variable number of nodes $(ht)$. After the new input $Xt$, the RNN's hidden state (latent memory) $h_t$ is updated with the nonlinear function $f$.

$$h_{t+1} = f(h_t, X_t) \qquad (6)$$

The capacity of $f$ in Equation (6) to deal with vanishing difficulties and exploding gradients, which are the most typical problems in the development and training of RNN, determines the value of $f$. Given the inputs $X_t, h_{t+1}$, the GRU updates for the time step $t$. To begin, we use the formula to calculate the update gate $z_t$ for the time step $t$:

When $Xt$ it is linked to a network unit, its own weight $W_z$ is multiplied. The same is true for $h_{t-1}$, which stores data from earlier $t-i$ units and is multiplied by their own weight $U_z$. Both results are combined, and the result is squashed between 0 and 1 using a sigmoid activation function. The update gate aids the model in determining how much previous data (from earlier time steps) should be passed on to the future. This is extremely useful since the model can duplicate all of the data from the past, eliminating the risk of disappearing gradients. The framework utilizes the reset gate to determine how much information from the past should be forgotten. We use the following formula to compute it in Equation (7),

$$r_t = \sigma(W_r x_t + U_r h_{t-1}) \qquad (7)$$

We create new memory content that will store relevant information from the past using the reset gate. It is calculated as follows in Equation (8),

$$h_t = \tanh(Wxt + r_t \odot Uh_{t-i}) \qquad (8)$$

This multiplies the input $x_t$ by weight *W* and the input $h_{t-1}$ by a weight U. Calculate the Hadamard (element-by-element) product between $r\_t$ and $Uh_{t-1}$. This will determine what time steps should be removed from the preceding ones. Add the results together and use the *tanh* nonlinear activation function. The network's final step is to calculate the $h_t$ vector, which contains information for the current unit and sends it down to the network. The update gate is required to accomplish this. It determines what should be collected from the present memory content $h't$ and what should be collected from the previous stages $h_{t-1}$. This is how it's done:

$$h_t = z_t \odot h_{t-1} + (1-z_t) \odot h'_t \tag{9}$$

This will execute element-wise multiplication on the updated gates $z_t$ and $h_{t-1}$, as well as $(1-z_t), h'_t$, and total the results. Below Fig.4 is a visual representation of GRU.

We want the representation of words to be such that the vectors for words with similar meanings or contexts are close to one other in the vector space. The word2vec algorithm, which turns a word into a vector, is likely the most used. A corpus of texts relating to the domain we are engaged in is used to train word2vec.

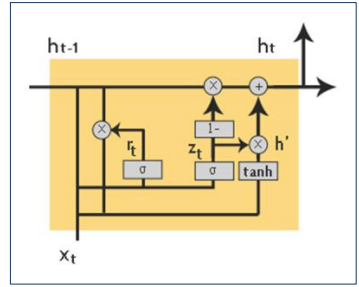

**Fig. 4.** Visual representation of Gated Recurrent Unit (GRU)

The word2vec algorithm, which turns a word into a vector, is likely the most used. A corpus of texts relating to the domain in which we are engaged is used to train word2vec. Word2vec trains a system that can anticipate the surrounding words of a target text in order to calculate the word vectors. Surrounding words are described as words that appear from both sides of a given word in a small context window of a particular size. These three sentences will serve as our corpus. (1) *This research paper is about deep learning and computer vision.* (2) *We love deep learning.* (3) *We love computer vision*.

The challenge is to predict the context terms *"learning," "and,"* and *"vision"* from the first phrase and *"we," "love,"* and *"vision"* from the last sentence given the word *"computer."* As a result, the goal of training is to maximize the log probabilities of these context terms given the word "computer." The formulation is in Equation (10),

$$\text{Objective} = \max \sum_{t=1-m \geq j \geq m}^{T} \sum \log P(E_{wt} + j | E_{wt}) \tag{10}$$

Where m is the size of the context window and t is the length of a corpus. The similarities or inner product between the context word vector and the center word vector represents $P(E_{wt} + j | E_{wt})$ when a word appears in context. When it is the central word, it has two vectors associated with it, denoted by *R and S*, accordingly. As a result, $P(E_{wt} + j | E_{wt})$ it is defined as:

$$\frac{e^{R^T_{Wt+j}}}{\sum_{i=1}^{M} e^{R^T_i s_{Wt}}} \tag{12}$$

The denominator is the normalization term, which compares the similarity of the central word vector to context vectors of every other word in the lexicon, resulting in a probability of one. The vector representation for a word is then chosen as the center vector, as with GRU.

### 3.3. Attention models

We apply the Bahdanau attention mechanism to better isolate the image content, which has been widely used to tackle the challenge of image categorization since it eliminates the need to process every pixel in an image. Instead of taking features from the entire image, the salient portion of the image is determined at each step and input into the RNN. The algorithm uses the image to create a focused view and predicts the term that is relevant to that location. On the basis of previously generated words, the location where attention is directed must be determined. Otherwise, new words formed within the region may be coherent, although not in the description generated. After that, proceed to the mechanism proposed by Bahadano. To begin, calculate the $e_{jt}$ score in Equation (12),

$$e_{jt} = f_{ATT}(E_{t-1}, h_j) \qquad (12)$$

$e_{jt}$ is a score that indicates how essential the $j_{th}$ pixel of an image is at every time step $t$ of the decoder. The prior state of the decoder is $S_{t-1}$, while the current state of the encoder is $h_j$. $f_{ATT}$ Is a basic feed-forward neural network that sums $S_{t-1}$ and $h_j$ from the fully connected layer, then passes it through a nonlinear function $tanh$ before returning to the fully connected layer.

$$e_{jt} = FC(tanh(FC(E_{t-1}) + FC(h_i))) \qquad (13)$$

We use softmax to get the probability distribution in Equation (15),

$$\alpha_{jt} = softmax(e_{jt}, \text{axis} = 1) \qquad (14)$$

Softmax usually is applied to the last axis, and however, since the shape of the score is
(batch_size, max_length, hidden_size), we want to apply it to the 1$^{st}$ axis. The maximum length of our input is max length. Softmax should be applied to that axis because we're aiming to allocate weight to each input. Now that we have the input, we must feed the decoder a weighted sum combination of the input.

$$c_t = \sum_{j=1}^{T} a_{jt} h_j \text{ suchthat } \sum_{j=1}^{T_x} a_{jt} = 1 \text{ and } a_{ij} \geq 0 \qquad (15)$$

The context vector (weighted total of the input) that will be sent to RNN is $c_t$.

$$E_t = RNN(S_{t-1}, e(y'_{t-1}), c_t) \qquad (16)$$

The previous state of the decoder is $S_{t-1}$, and the last predicted word is $e(y'_{t-1})$.

### 4. Experimental Setup and Results

The experimental setup aimed to evaluate the effectiveness of a proposed framework for image captioning using the MSCOCO and Flickr30k dataset. These datasets are widely used benchmarks for image captioning and consist of many images with corresponding natural language descriptions. To evaluate the effectiveness of the proposed framework, the authors likely used metrics such as BLEU, ROUGE, CIDER, and METEOR to compare the generated captions to the ground truth captions. These metrics measure the similarity between the generated and ground truth captions based on n-gram overlap, recall, and other factors.

### 4.1. Datasets

*MSCOCO:* MSCOCO dataset [9], which contains 82,783, 40,504, and 40,775 images for training, validation, and testing, is the most significant benchmark dataset for the image captioning task. Because most images feature many objects in the context of complicated situations, this dataset is difficult to analyze. Each image in this dataset has five captions with different ground truths that have been annotated by humans, as seen in Fig. 4. By utilizing the same data split as in [23] for offline assessment, which consists of 5,000 photos for validation, 5,000 images for the test, and 113,287 images for training.

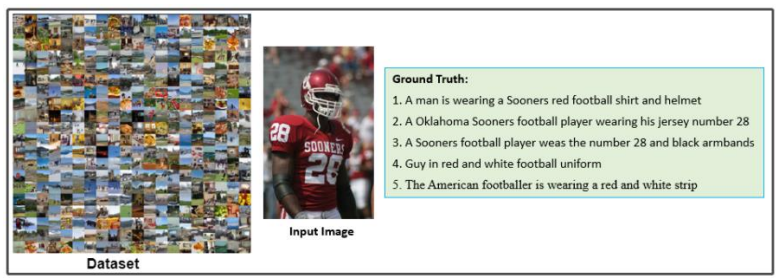

**Fig 4.** Representation of dataset.

After that, combine the testing set with the training set to create a more extensive training set for online evaluation on the MSCOCO evaluation server. Then, remove all non-alphabetic characters from the captions, convert all letters to lowercase, then tokenize the captions with white space. As a result, a vocabulary of 9,487 terms has been created.

*Flickr30K:* Flickr30K is an automated image interpretation and grounded language comprehension dataset. It comprises 30K Flickr images and has 158K captions generated by personal annotators. It doesn't have a specified division of images for analysis and justification of instruction. For preparation, measurement, and evaluation, investigators may select their choice numbers. The dataset also includes typical object detectors, a color classifier, and a tendency against more significant object collection. The Flickr30k dataset contains 31,000 images, each with five captions. Table 1 compares the splitting ratios of various datasets briefly.

Table1: splitting ratios of MSCOCO and Flicker30k datasets

| Dataset | Train Split | Validation Split | Test Split |
|---|---|---|---|
| MSCOCO | 113,287 | 5,000 | 40,504 |
| Flicker30k | 29,000 | 1,014 | 1,000 |

### 4.2. Evaluation measurements

Four commonly used evaluation metrics, namely BLEU 1-4 (Bilingual Evaluation Understudy) [32], Meteor (Metric for Evaluation of Translation with Explicit ORdering) [33], Rouge-L (Recall-Oriented Understudy for Gisting Evaluation - Longest Common Subsequence) [34], and CIDEr (Consensus-based Image Description Evaluation) [35], are utilized to evaluate the quality of generated sentences utilizing the publicly accessible MSCOCO tools to quantitatively analyze the performance of our proposed methods. All of these metrics are used to compare the consistency of n-grams in generated and reference sentences. To make accurate comparisons with existing image captioning approaches.

## 4.3. Quantitative Results

Demonstrating the quantitative results in this section shows the efficacy of the proposed strategy. The suggested technique is compared to seven state-of-the-art models on MSCOCO and Flickr30k in a multi-comparison, as shown in Table 2 and Table 4, respectively. In the first-time step of the LSTM-based language model, NIC injects image features derived from the fully connected layer of a deep CNN. The results presented are cited directly [36]. Soft-Att selects some regional representations from the deep CNN's final convolutional layer and uses an LSTM-based language model to decode each word at each time step, depending on the representations selected [31]. MSM incorporates inter-attribute correlations into a multiple-instance learning approach and investigates several techniques of injecting detected characteristics and image representations into an LSTM-based language framework [2]. Attribute-driven using CNN-RNN architecture and the visual attention method for attribute detector, attention model the co-occurrence dependencies among attributes [37]. NBT architecture for visually grounded image captioning generates free-form natural language descriptions while localizing things in the image [38].

The relationship between the objects/regions in an image is modeled with a GNN in visual context-aware attention, which considers the visual relationship between areas of interest for improved representation of the visual content in the image [39]. The elegant views of what kind of visual relationships could be built between objects and how to nicely leverage such visual relationships to learn more informative and relation-aware region representations come from GCN-use LSTMs of visual relationships for enriching region-level models and eventually enhancing and lead to the elegant views of what kind of visual relationships could be built between objects and how to nicely leverage such visual relationships to learn more informative and relation-aware region representations [40]. The bold number represents the top results for that measure, whereas those with a dash (--) are unavailable.

**Table 2.** Comparison of the proposed GRU attention-based model with various baseline approaches on the MS COCO dataset. The dash (–) is unavailable.

| MODEL | BLEU-1 | BLEU-2 | BLEU-3 | BLEU-4 | Rouge | CIDER | METEOR |
|---|---|---|---|---|---|---|---|
| Google NIC [36] | 0.67 | 0.45 | 0.30 | 0.20 | -- | -- | -- |
| Soft Attention [31] | 0.71 | 0.49 | 0.34 | 0.24 | -- | -- | 0.24 |
| MSM [2] | 0.73 | 0.57 | .043 | 0.33 | 0.54 | 1.02 | 0.25 |
| Attribute-driven Att [37] | 0.74 | 0.56 | 0.44 | -- | 0.55 | 1.104 | -- |
| NBT [38] | 0.75 | -- | 0.34 | -- | -- | 1.107 | 0.27 |
| Context-aware Att [39] | 0.76 | 0.60 | 0.46 | 0.36 | 0.56 | 1.103 | 0.28 |
| GCN-LSTM [40] | 0.77 | -- | -- | 0.36 | 0.57 | 1.107 | 0.28 |

**Table 3.** State-of-the-art approaches were compared to the performance of different pre-trained models for encoder architecture on the MSCOCO dataset. The bold number represents the top results.

| MODEL | BLEU-1 | BLEU-2 | BLEU-3 | BLEU-4 | Rouge | CIDER | METEOR |
|---|---|---|---|---|---|---|---|
| Inception V3 | **0.78** | **0.57** | **0.44** | 0.36 | **0.59** | 1.105 | 0.27 |
| VGG16 | 0.74 | **0.57** | **0.44** | 0.33 | 0.56 | **1.109** | 0.26 |
| DenseNet169 | 0.74 | 0.56 | 0.43 | 0.36 | **0.58** | 1.103 | 0.27 |
| ResNet101 | 0.75 | 0.56 | **0.44** | **0.37** | **0.59** | 1.104 | **0.29** |

**Table 4.** Comparison of the proposed network with existing image illustration techniques on the Filker30k dataset. The dash (-) is not available.

| Model | BLEU-1 | BLEU-2 | BLEU-3 | BLEU-4 | METEOR | CIDEr | ROUGE |
|---|---|---|---|---|---|---|---|
| Google-NIC [36] | 0.66 | 0.42 | 0.27 | 0.18 | -- | -- | -- |
| m-GRU[41] | 0.66 | 0.40 | 0.28 | 0.20 | 0.29 | 0.48 | -- |
| phi-LSTM[42] | 0.64 | 0.45 | 0.31 | 0.21 | 0.19 | 0.45 | 0.44 |
| Soft-Att [31] | 0.66 | 0.43 | 0.28 | 0.19 | 0.18 | -- | -- |

**Table 5.** For the Filker30k dataset, cutting-edge techniques were compared to the performance of various pre-trained models for encoder design. The top results are represented by the bold number.

| MODEL | BLEU-1 | BLEU-2 | BLEU-3 | BLEU-4 | Rouge | CIDER | METEOR |
|---|---|---|---|---|---|---|---|
| Inception V3 | **0.70** | **0.47** | **0.36** | 0.26 | **0.47** | 0.49 | 0.28 |
| VGG16 | 0.67 | 0.46 | **0.36** | 0.24 | 0.44 | **0.47** | **0.29** |
| DenseNet169 | 0.65 | 0.45 | 0.34 | **0.27** | 0.46 | **0.47** | 0.27 |
| ResNet101 | 0.67 | 0.46 | **0.36** | 0.26 | 0.46 | 0.46 | 0.28 |

To compare the performance of our proposed model with other state-of-the-art approaches, we utilized different pre-trained models for encoder architecture by evaluating the quantitative analysis of the presented framework's four metrics: BLEU, Rouge, CIDER, and Meteor. Table 3 shows the results for the MSCOCO dataset, while Table 5 displays the results for the Fliker30k dataset. The top performers for each metric are indicated by *bold* numbers. Inception outperformed the other models, followed by ResNet10. While the overall results are satisfactory, we acknowledge using 113,287 images for training from the MSCOCO dataset and 29,000 from the Filker30k dataset. Figure 5 (a) illustrates the loss plot for four pre-trained networks. This could be useful in evaluating the training process and comparing the performance of different pre-trained networks. Figures 5 (b) and 6 dispute the visual representation for the performance of proposed GRU attention-based models.

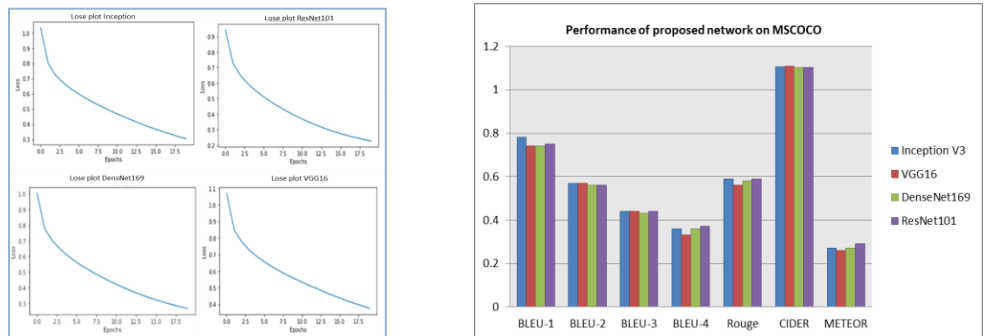

**Fig 5 (a).** Loss Plot of models pre-trained with different networks  **(b)** Statistical evaluation score of different models on the MSCOCO dataset

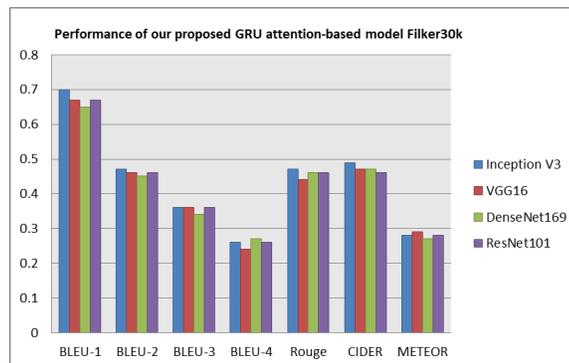

**Fig.6.** Statistical evaluation score of different models on the Filker30K dataset

### 4.4. Qualitative results

The qualitative results of the proposed model are intriguing as it generates relevant and grammatically correct captions for various images. Figure 7 demonstrates positive outcomes on the MSCOCO dataset, while Figure 8 illustrates the caption generation on the Flickr30k dataset. Although most captions are informative, there are instances where the generated caption describes a situation different from the image or is entirely incomprehensible. These errors can be attributed to two tasks performed by the system, namely, image recognition and text generation. The former can be labeled as a failure in image recognition, while the latter can be attributed to a failure in text generation. Figure 9 presents several examples of such failures.

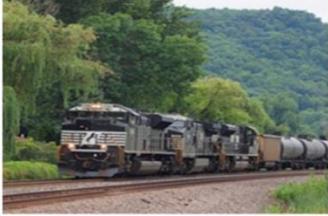

**Fig. 7.** Example captions from the conventional approach with our GRU-based encoderdecoder model, as well as their ground truth captions for MSCOCO dataset.

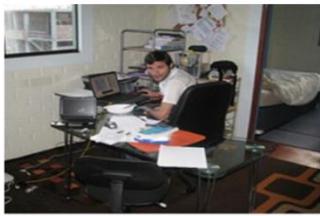

**Fig. 8.** Image-to-text retrieval results.from the conventional approach with our GRU-based encoder-decoder model, as well as their ground truth captions for the Filker30K dataset.

We added relevant test experiments and verified the effectiveness of the developed framework, increasing the automatic synthesis results of captioning obtained from the processed randomly sampled images in Figure 11 (supplementary data) (i.e., image samples are not in the training set and validation set) that shown that the developed framework is effective at generating accurate caption for a wide range of images. The network's tasks cannot be distinguished, but they can be labeled as failures in image recognition and text generation, respectively. Figure 9 provides examples of these failures, where the proposed captioning method generates inaccurate captions due to errors in detecting and recognizing objects and their contextual relationships. As a result, the process fails to extract the intended semantics from specific images.

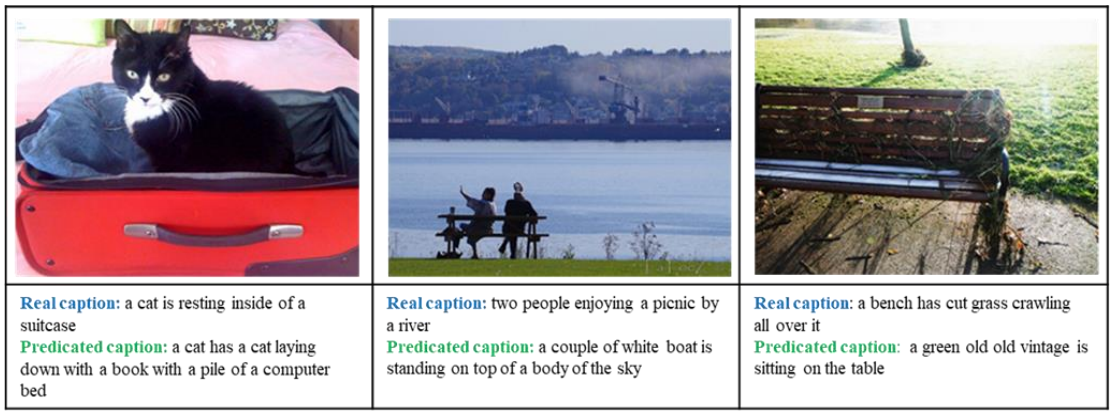

**Fig 9.** The model predicts poor captions in this case.

The example output of the attention network architecture is illustrated in Figure 9. The title of the figure is informative, and the attention of the network seems to be concentrated on the word "teddy bears," which is located in the section of the image that is colored like a teddy bear.

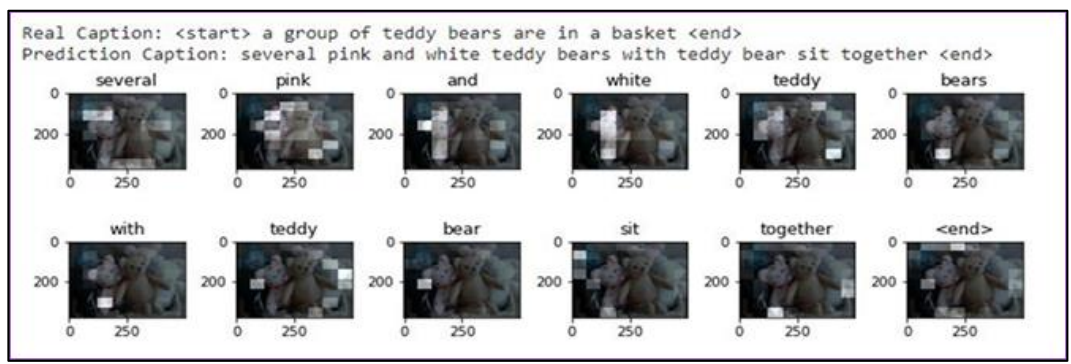

**Fig. 10.** Attention is drawn to different regions of the image. The attention weights in white spots are higher

## 5. Conclusions and Future Scope

In this research work, we proposed a single joint model for automatic image captioning using a combination of CNN and GRU with an attention network. The suggested framework outperforms pre-trained CNNs and significantly improves caption quality, as demonstrated by both qualitative and quantitative experiments using the MS COCO dataset. Presented findings show that the Bahdanau attention model combined with GRU can effectively focus on specific regions of the image and enhance the overall performance of the model. The proposed approach can help bridge the gap between computer vision and natural language processing and extend caption generation into specific domains. Future research can explore various directions to enhance the performance of image captioning models. At first possible direction is to investigate the use of more advanced attention mechanisms, such as the Transformer-based models, to improve further the model's ability to focus on specific regions of the image. Furthermore, explore the use of alternative language models, such as the LSTM, to assess their impact on caption quality. Additionally, incorporating external knowledge sources, such as common sense reasoning, could be beneficial to generate captions that are more informative and accurate. Finally, including a feedback loop into the model to iteratively refine the captions could be explored as a promising direction for future research.

**Availability of Data and Systems**

In all of the experiments, the NVIDIA GPU GTX-1070 was employed. This model was created on a GPU with 64GB of memory and an Intel i9 9900k processor. The development process was completed using PyTorch..

**Conflicts of Interest:** There are no conflicts of interest to report by the authors. The study design, data collection, analysis and interpretation, manuscript writing, and publication decisions were not influenced by the funders..

**Acknowledgment:** The Fundamental Research Funds support this research for the Central Universities. (Grant no.WK2350000002).

**Author Contributions:** Rashid Khan: Concept and design,data acquisition,analysis, interpretation of data and draft writing. Haseeb Hasan: Writing, rewiew and editing. Bingding Huang: Technical administrative draft proofread. Zhongfu Ye: Investigation, review, editing, and supervision.